\documentclass[sigconf]{acmart}

\usepackage{booktabs}
\usepackage{microtype}
\usepackage{comment}
\usepackage{algorithm}
\usepackage{algorithmic}
\usepackage{tikz}
\usetikzlibrary{arrows.meta, positioning, fit, shapes.geometric}
\usepackage{soul}
\usepackage{subcaption}
\usepackage{multirow}
\usepackage{listings}
\usepackage{longtable}
\usepackage{colortbl}
\usepackage[most]{tcolorbox}
\usepackage{dsfont}
\usepackage{hyperref}
\hypersetup{
    colorlinks=true,
    linkcolor=blue,
    citecolor=blue,
    urlcolor=blue
}

\acmConference[RecSys '26]{20th ACM Conference on Recommender Systems}{October 2026}{Minneapolis, USA}

\author{Akshay Kekuda}
\affiliation{\institution{Walmart Global Tech}\city{Bellevue}\state{WA}\country{USA}}
\email{akshay.kekuda@walmart.com}

\author{Shreeranjani Srirangamsridharan}
\affiliation{\institution{Walmart Global Tech}\city{Sunnyvale}\state{CA}\country{USA}}
\email{shreeranjani.srirangamsridharan@walmart.com}

\author{Ishan Bhatt}
\affiliation{\institution{Walmart Global Tech}\city{Sunnyvale}\state{CA}\country{USA}}
\email{ishan.bhatt@walmart.com}

\author{Yanan Cao}
\affiliation{\institution{Walmart Global Tech}\city{Sunnyvale}\state{CA}\country{USA}}
\email{yanan.cao@walmart.com}

\author{Sinduja Subramaniam}
\affiliation{\institution{Walmart Global Tech}\city{Sunnyvale}\state{CA}\country{USA}}
\email{sinduja.subramaniam@walmart.com}

\author{Evren Korpeoglu}
\affiliation{\institution{Walmart Global Tech}\city{Sunnyvale}\state{CA}\country{USA}}
\email{evren.korpeoglu@walmart.com}

\author{Kaushiki Nag}
\affiliation{\institution{Walmart Global Tech}\city{Sunnyvale}\state{CA}\country{USA}}
\email{kaushiki.nag@walmart.com}

\author{Kannan Achan}
\affiliation{\institution{Walmart Global Tech}\city{Sunnyvale}\state{CA}\country{USA}}
\email{kannan.achan@walmart.com}

\setcopyright{none}
\renewcommand\footnotetextcopyrightpermission[1]{}
\begin{document}

\title{Timing-Aware Repurchase Prediction for Web-Scale E-Commerce: Survival Models for Multi-Surface Grocery Recommendation}

\begin{abstract}
Repurchase recommenders in e-commerce are commonly framed as a binary question asking \emph{will this customer buy this item within $W$ days}, a formulation that in our production setting requires a separately trained model for every horizon of interest.
We replace this stack with survival models that predict
\emph{time-to-repurchase} directly, and evaluate them on millions of customers from a major grocery e-commerce
platform across more than thirty ablation configurations.
Our study makes three contributions.
First, an empirical hazard analysis reveals a slightly
\emph{decreasing} marginal hazard ($\hat{k} \approx 0.9$). This differs from the the common intuition that grocery items become more likely to be
repurchased the longer it has been since the last purchase (an
\emph{increasing} hazard with $k > 1$).
Log-Normal achieves the best marginal fit ($R^2{=}0.998$) and the
best ranking, despite Weibull providing the best fit to conditional
residuals, revealing an apparent discrepancy that we analyze in detail.
Second, a single Accelerated Failure Time (AFT) model \emph{replaces
three per-horizon binary classifiers}, matching or exceeding each at
its own horizon while using roughly $3\times$ fewer total trees.
Feature importance reshuffles under the survival objective:
channel-cadence and recency signals rise sharply while aggregate
frequency counts fall.
Third, a 4-parameter parametric calibration maps raw survival
cumulative distribution functions (CDFs) to per-horizon
probabilities with zero cross-horizon monotonicity violations.
Calibration quality varies by an order of magnitude across the AFT
family even though ranking does not:
Exponential AFT (Weibull $k{=}1$) achieves expected calibration
error (ECE) $\approx 10^{-4}$, roughly $10\times$ lower than
Log-Normal, while ranking metrics agree within $0.3\%$ relative.
We therefore adopt Exponential AFT for probability-consuming
surfaces and Log-Normal for pure ranking, exposing a principled
calibration--ranking trade-off within a single AFT family.
\end{abstract}


\maketitle

\section{Introduction}
\label{sec:intro}

Online grocery platforms serve multiple recommendation surfaces
with different time horizons.
On the platform we study, for example, a 7-day shopping list appears
on the homepage, a 14-day restocking carousel in the mobile app, and
a 30-day pantry planner in email.
Each surface needs a ranked list of previously-purchased items, but
the underlying question is not really about  \emph{what} the customer will buy. It is about \emph{when}.
Milk repurchased weekly, laundry detergent monthly, and baking soda
every few months all sit in the same candidate set, yet on wildly
different clocks.

The standard production approach ignores those clocks.
An XGBoost~\cite{chen2016xgboost} \texttt{binary:logistic}
classifier asks ``will this customer buy this item within $W$
days?'', which means an item due in two days and one due in three
weeks can receive the same score.
When the platform needs to populate three surfaces at different
horizons, the pragmatic but wasteful solution is to train three
models, one per horizon, tripling training and serving cost while
discarding all timing structure in the process.

Survival analysis~\cite{cox1972regression,kalbfleisch2002statistical}
offers a natural alternative: instead of a binary yes/no, we predict
\emph{time-to-repurchase}.
We study two complementary formulations.
\emph{Accelerated Failure Time} (AFT) models~\cite{wei1992accelerated}
predict a scalar time-to-repurchase per (customer, item) pair,
yielding an imminence-ordered ranking that serves every horizon from
a single model.
\emph{Discrete-Time} (DT) hazard models~\cite{allison1982discrete,singer2003applied}
expand each pair into interval-specific rows, producing calibrated
per-horizon CDFs and, in principle, a \emph{different ranked list per
horizon} from a single model.

Our contributions:
\begin{enumerate}
  \item \textbf{Data-driven distribution selection.}
    An empirical hazard analysis on tens of millions of
    (customer, item) pairs yields a maximum-likelihood estimate
    (MLE) of the shape parameter
    $\hat{k}_\text{MLE} = 0.911$, a slightly
    \emph{decreasing} hazard, which contrasts with the common
    intuition that repurchase hazards should be increasing.
    We compare Weibull, Log-Normal, and Logistic AFT distributions
    across marginal fit (via Kaplan--Meier quantile--quantile
    regression), conditional residual fit, and end-to-end ranking.
    Log-Normal achieves the best marginal fit ($R^2{=}0.998$) and
    the best ranking metrics, despite Weibull providing the best fit
    to conditional residuals, revealing an apparent discrepancy that we analyze in detail
    (\S\ref{sec:empirical_hazard}, \S\ref{sec:shape_ablation}).

  \item \textbf{One model replaces three.}
    A single AFT model matches or exceeds the best
    per-horizon binary classifier at every horizon while using
    roughly $3\times$ fewer total trees across horizons.
    The survival objective reshuffles feature importance:
    channel-cadence and recency features rise sharply while
    aggregate frequency counts drop
    (\S\ref{sec:main_results}, \S\ref{sec:feature_importance}).

  \item \textbf{Calibration separates the AFT family.}
    A 4-parameter parametric calibration (shape $a$ plus per-horizon
    intercepts $b_t$) maps raw survival CDFs to per-horizon
    probabilities with zero cross-horizon monotonicity violations.
    Although ranking is nearly invariant across AFT distributions,
    calibration quality varies by an order of magnitude:
    Exponential AFT (Weibull $k{=}1$) reaches ECE $\approx 10^{-4}$
    with near-perfect sanity ratios ($1.00$--$1.01$),
    while Log-Normal, despite winning marginal fit and
    ranking, trails by $10\times$ on ECE due to its bell-shaped
    hazard mismatching the empirical monotone-decreasing hazard.
    We recommend Exponential for probability-consuming surfaces and
    Log-Normal for pure ranking
    (\S\ref{sec:calibration}).
\end{enumerate}

\section{Problem Setting}
\label{sec:problem}

\paragraph{Task.}
Given a customer's purchase history on a web-scale grocery
e-commerce platform, rank previously-purchased items by likelihood
and timing of future repurchase.
The output populates ``buy again'' surfaces across multiple
web and mobile applications, each with a different time horizon.

\paragraph{Data.}
Our dataset consists of tens of millions of (customer, item) pairs
drawn from a large, proprietary grocery e-commerce platform.
Each pair is labeled with the observed time-to-repurchase in days if
a repurchase occurred within a 30-day window, and is right-censored
otherwise.
Training uses a single snapshot date; evaluation uses the same date
with a held-out customer partition.

\paragraph{Evaluation protocol.}
Our primary metric is \emph{precision-at-variable-$k$}
(P@$h$): for each customer, we retrieve $k$ items where $k$ equals
the number of items actually purchased within the $h$-day window.
This makes precision and recall numerically identical, evaluating
ranking quality at each customer's natural basket size.
We evaluate at three horizons: $h \in \{7, 14, 28\}$ days.
Secondary metrics include normalized discounted cumulative gain
(NDCG), consumable-item precision, perishable-item precision, and
fixed-$k$ precision at $k \in \{4, 8, 16\}$.

\paragraph{Baseline.}
Our production system trains an XGBoost classifier with \mbox{\texttt{binary:logistic}} (800 trees, depth 6, learning rate (LR) 0.3, 84 features)
on 14-day repurchase labels.
For 7-day or 30-day recommendations, separate models are trained with
horizon-matched labels using the same architecture and features.
This yields three independent models $F_\text{7d}$, $F_\text{14d}$,
$F_\text{30d}$.
Each is evaluated on all horizons; the matched-horizon result
(e.g., $F_\text{14d}$ at 14 days) represents each model's ceiling.

\paragraph{Features.}
All models share 84 features:
order frequency by fulfillment channel and payment method,
recency signals (days since last purchase, days-since-last-$N$-order),
inter-purchase interval (IPI) statistics at item and product-type
level (personal median, population median, percentiles),
item availability flags,
customer tenure features,
basket composition signals,
and previous-order presence flags.
The best AFT model uses the same production feature set;
the DT model adds interval-relative features (\S\ref{sec:dt}).

\section{Related Work}
\label{sec:related}

\paragraph{Next-basket and repurchase ranking.}
Predicting what a customer will buy next is most often studied as
next-basket recommendation (NBR).
The literature spans Markov-chain factorization~\cite{fpmc},
recurrent basket encoders~\cite{yu2016dream}, attention- and
set-based encoders~\cite{sun2019bert4rec,hu2019sets2sets,le2019beacon,yu2020dnntsp},
and frequency-based nearest-neighbor methods such as
TIFU-KNN~\cite{hu2020tifuknn}.
A parallel line of work isolates the \emph{repeat} portion of the
basket, which is precisely what buy-again surfaces consume:
RepeatNet~\cite{ren2019repeatnet} disentangles repeat from explore
intent, ReCANet~\cite{ariannezhad2022recanet} models per-item repeat
consumption for grocery next-basket recommendation, Wan and
McAuley~\cite{wan2018representing} model basket-level loyalty and
complementarity, and Bhagat et al.~\cite{bhagat2018buyitagain}
formulate repeat purchase as binary classification.
Closest to our own setting, CASE~\cite{case} encodes each item's
purchase history as a calendar-time signal and models cross-candidate
dependencies with induced set attention.
Reproducibility studies in this area repeatedly find that simple,
well-tuned baselines remain competitive with substantially more
elaborate neural models on grocery data~\cite{li2023realitycheck}.

\paragraph{Survival analysis in recommendation.}
What all of the models above share is that they answer \emph{what}
will be rebought and, at best, how likely; none predicts \emph{when},
which is the quantity a multi-horizon repurchase stack actually needs.
Survival analysis targets timing directly.
Survival models have been applied to user engagement in online
experiments~\cite{chandar2022survival} and to user return-time
modeling~\cite{kapoor2015just}, but their use for \emph{item-level
repurchase ranking} at web scale is underexplored.
Our work shows that XGBoost AFT is a drop-in replacement for binary
logistic without architecture changes.

\paragraph{Time-aware and multi-horizon recommendation.}
Session-based models (GRU4Rec~\cite{hidasi2016session},
Time-LSTM~\cite{zhu2017next}) model temporal dynamics but require
sequence architectures and do not output calibrated per-horizon
probabilities.
Multi-horizon prediction is typically approached either by training
one model per horizon or via multi-task
learning~\cite{ma2018modeling}.
Person-period expansion (\S\ref{sec:dt}) provides an alternative:
calibrated per-horizon probabilities via data augmentation rather than
architectural modification.

\paragraph{Discrete-time survival.}
The person-period expansion~\cite{efron1988logistic,singer2003applied} is
well established in biostatistics but rarely applied to web-scale
recommendation.
We show that it pairs naturally with gradient-boosted trees on
expanded datasets containing hundreds of millions of interval rows.

\paragraph{Positioning: our baseline is the deployed model.}
The repurchase ranker in production on our platform is a tabular
feature set scored by gradient-boosted trees (\S\ref{sec:problem}),
not a neural NBR architecture.
This is a deliberate engineering choice rather than an oversight.
In internal benchmarking on our own traffic, tuned tabular GBDT
rankers were competitive with neural NBR models while training and
serving at a small fraction of the cost.
They also remained directly explainable through feature
attributions, which is critical for a surface that merchandising and
category teams must be able to interrogate.
This is consistent with a broader body of evidence that carefully
tuned baselines are difficult to beat in
recommendation~\cite{dacrema2019,rendle2020neural} and that tree
ensembles remain strong on heterogeneous tabular
data~\cite{shwartzziv2022tabular,grinsztajn2022why}.
Accordingly, this paper does not attempt a head-to-head comparison
against neural NBR systems, and we make no claim of state-of-the-art
accuracy against them.
We compare instead against the model that is actually deployed, under
its own feature set, training budget, and evaluation protocol, and
our claim is scoped to match: changing the \emph{objective} from
per-horizon binary classification to time-to-event, while holding
the tabular representation and the tree ensemble fixed, improves
multi-horizon ranking and collapses three production models into one.

\section{Survival Models}
\label{sec:models}

\subsection{Accelerated Failure Time (AFT)}
\label{sec:aft}

We train XGBoost with the \texttt{survival:aft} objective.
For a (customer, item) pair with feature vector $\mathbf{x}$, the
model predicts an expected time-to-repurchase
$\hat{\lambda}(\mathbf{x}) > 0$.
All three AFT distributions share the same ranking rule:
items are ordered by predicted time $\hat{\lambda} = \exp(f(\mathbf{x}))$;
only the training loss gradient differs.
We compare Weibull, Log-Normal, and Logistic in
\S\ref{sec:shape_ablation}.

\paragraph{Survival, CDF, and hazard.}
Let $\hat{\lambda} = \exp(f(\mathbf{x}))$ denote the predicted
time-to-repurchase.

\vspace{4pt}
\noindent\textit{Weibull} (shape $k$, scale $\hat{\lambda}$):
\begin{align}
S(t) &= \exp\!\bigl(-(t/\hat{\lambda})^k\bigr), \quad
F(t) = 1 - S(t) \label{eq:surv} \\
h(t) &= \tfrac{k}{\hat{\lambda}}\,(t/\hat{\lambda})^{k-1}
\label{eq:hazard}
\end{align}
For $k > 1$ the hazard increases (items become ``overdue'');
for $k < 1$ it decreases; $k{=}1$ is memoryless exponential.
Our empirical analysis (\S\ref{sec:empirical_hazard}) finds
$\hat{k}_\text{MLE} = 0.911$.

\noindent\textit{Log-Normal} (location $\mu{=}\ln\hat{\lambda}$, scale $\sigma$):
\begin{align}
F(t) &= \Phi\!\bigl((\ln t - \mu)/\sigma\bigr), \quad
h(t) = \frac{f(t)}{S(t)}
\label{eq:lognormal}
\end{align}
where $\Phi$ is the standard normal CDF.
The hazard is non-monotone (rises then falls).

\noindent\textit{Logistic} (location $\mu{=}\ln\hat{\lambda}$, scale $s$):
\begin{align}
F(t) &= \sigma\!\bigl((\ln t - \mu)/s\bigr), \quad
h(t) = \frac{f(t)}{S(t)}
\label{eq:logistic}
\end{align}
where $\sigma(\cdot)$ is the sigmoid function.
The hazard is also non-monotone.

\paragraph{Peak repurchase day.}
Each distribution yields a closed-form density mode (most likely
repurchase day):
\begin{align}
t^*_\text{Weibull} &= \hat{\lambda} \cdot \bigl((k{-}1)/k\bigr)^{1/k}
  \quad (k > 1; \text{mode}{=}0 \text{ if } k \le 1)
\label{eq:mode_w} \\
t^*_\text{LogN} &= \exp(\mu - \sigma^2)
\label{eq:mode_ln} \\
t^*_\text{Logistic} &= \exp(\mu) \cdot \bigl((1{-}s)/(1{+}s)\bigr)^{s}
  \quad (s < 1)
\label{eq:mode_lg}
\end{align}
Log-Normal always has a finite mode, making it directly
interpretable as a ``peak repurchase day'' even when the Weibull
mode collapses to $t{=}0$ (as it does for our empirical
$k < 1$).
Ranking by $-\hat{\lambda}$ remains valid regardless of
distribution.

\paragraph{Ranking score.}
Items are ranked by $-\hat{\lambda}(\mathbf{x})$: smaller scale
(sooner expected repurchase) ranks first.
This is a drop-in replacement for the binary classifier, using same
inference pipeline with the same latency.

\paragraph{Censoring.}
Items without a repurchase within 30 days are right-censored with
lower bound equal to elapsed days since last purchase and upper bound
$+\infty$.
Preserving all censored observations is critical: three experiments
that dropped censored items with high lapsed-days ($>$180d)
collapsed to concordance 0.55--0.65 vs.\ 0.91+, even with heavy
regularization (\S\ref{sec:aft_ablations}).

\paragraph{Ranking invariance.}
For any single-output AFT model (Weibull, Log-Normal, or Logistic),
item rankings are identical across horizons.
$F(h \mid \mathbf{x})$ is monotone decreasing in $\hat{\lambda}$
for fixed $h$ and distribution parameters, so:
\begin{equation}
\operatorname{argsort}\bigl(F(h_1 \mid \mathbf{x})\bigr)
= \operatorname{argsort}\bigl(F(h_2 \mid \mathbf{x})\bigr)
\quad \forall\, h_1, h_2 > 0
\end{equation}
A single model serves all horizons.
Per-horizon \emph{probabilities} are obtained via calibration
(\S\ref{sec:calibration}), but the ranking is horizon-agnostic.

\subsection{Discrete-Time Hazard (DT)}
\label{sec:dt}

The discrete-time hazard model~\cite{allison1982discrete} frames
repurchase as a sequence of interval-specific binary events.
We define three intervals: $[0, 7)$, $[7, 14)$, $[14, 30)$ days.
Each (customer, item) pair is expanded into one row per interval
(\emph{person-period expansion}), yielding $\approx 2.9{\times}$
expansion.
For uncensored pairs, the row corresponding to the event interval
receives label 1; preceding rows receive label 0.
Censored pairs receive label 0 for all rows.
We retain 30\% of censored items via negative sampling.

A standard \texttt{binary:logistic} XGBoost model is trained on the
expanded dataset with the interval index as a categorical feature.

\paragraph{Interval-relative features.}
Six interval-relative features simulate the customer's state at each
interval boundary, using the elapsed time since last purchase
($\text{lapsed}$) and the interval endpoint offset ($d_\text{end}$):

\vspace{2pt}
\begin{center}
\footnotesize
\begin{tabular}{lp{4.6cm}}
\toprule
Feature & Definition \\
\midrule
Lapsed-at-end
  & $\text{lapsed} + d_\text{end}$ \\
IPI gap at end
  & $\text{IPI}_\text{personal} - (\text{lapsed} + d_\text{end})$ \\
Fraction of IPI elapsed
  & $(\text{lapsed} + d_\text{end}) / \text{IPI}_\text{personal}$ \\
Population IPI gap
  & $\text{IPI}_\text{pop} - (\text{lapsed} + d_\text{end})$ \\
Lower-bound gap
  & $p_5 - (\text{lapsed} + d_\text{end})$ \\
Upper-bound gap
  & $p_{95} - (\text{lapsed} + d_\text{end})$ \\
\bottomrule
\end{tabular}
\end{center}
\vspace{2pt}

\paragraph{Inference.}
Interval hazards $h_j = P(\text{event in } j \mid
\text{survived to } j)$ yield survival CDFs:
\begin{align}
F(7) &= h_0 \nonumber \\
F(14) &= 1-(1-h_0)(1-h_1) \nonumber \\
F(30) &= 1-\prod_{j=0}^{2}(1-h_j) \nonumber
\end{align}
Ranking at horizon $h$ uses $F(h)$, producing a
\emph{different ranked list per horizon} from a single model.

\section{Experiments}
\label{sec:experiments}

\subsection{AFT Ablations}
\label{sec:aft_ablations}

We trained more than thirty AFT configurations varying distribution,
shape, depth, regularization, and features.
Two findings stand out: the distributional assumption can be estimated from data rather than treated as a free choice, and the marginal distribution with the best fit, Log-Normal, also produces the best ranking.

\subsubsection{\textbf{Empirical Hazard Analysis}}
\label{sec:empirical_hazard}

\begin{figure}[t]
  \centering
  \includegraphics[width=\columnwidth]{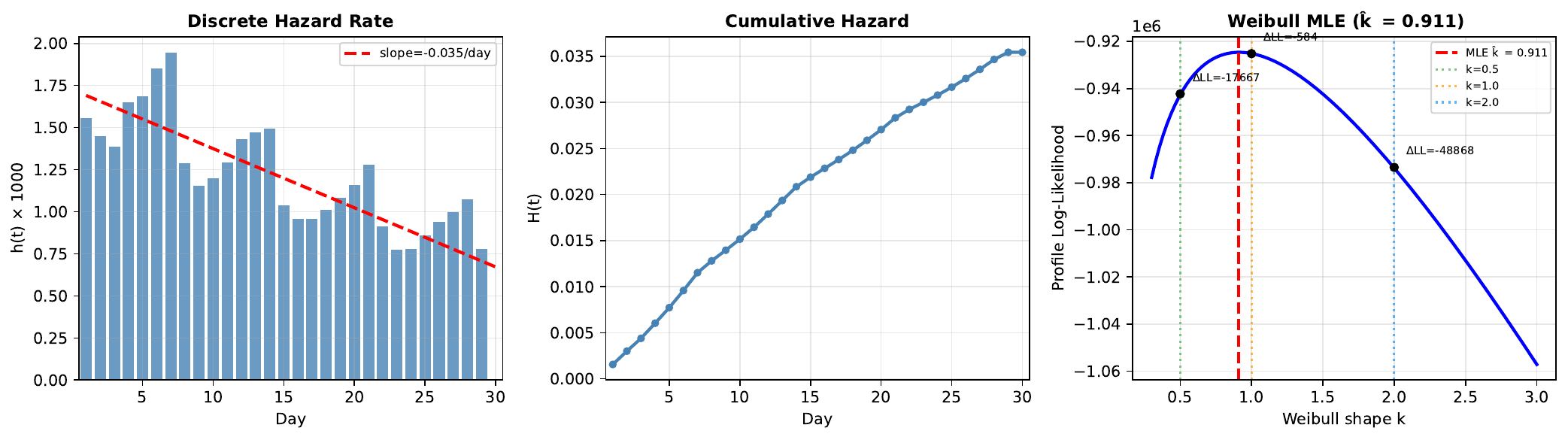}
  \caption{Empirical hazard analysis on a large
  customer--item sample.
  \textbf{Left}: discrete hazard rate $h(t)$ with decreasing linear trend
  (slope $= -0.035$/day) and weekly periodicity.
  \textbf{Center}: cumulative hazard $H(t)$.
  \textbf{Right}: Weibull profile log-likelihood yields
  $\hat{k}_\text{MLE}{=}0.911$; $k{=}1$ is nearest to the optimum.}
  \label{fig:hazard}
\end{figure}

We estimate the true hazard shape from a large sample of
(customer, item) pairs observed over a 30-day window.
The discrete hazard rate $h(t)$ shows a clearly \emph{decreasing}
trend (slope $= -0.035$/day): items are more likely to be
repurchased early in the window than late
(Figure~\ref{fig:hazard}).

\begin{figure}[t]
  \centering
  \includegraphics[width=\columnwidth]{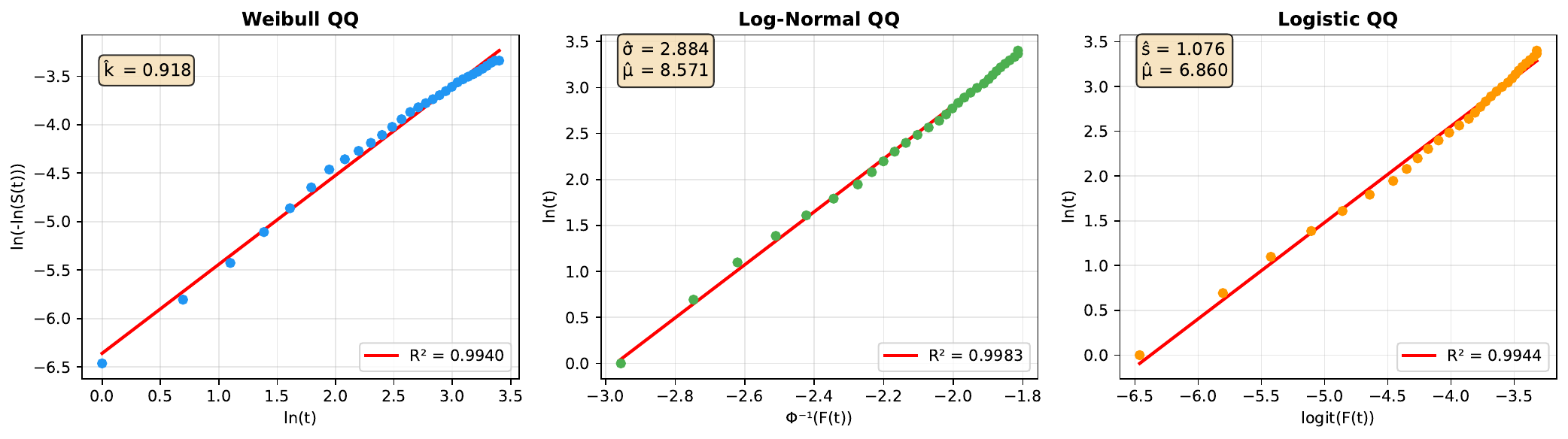}
  \caption{Marginal distribution comparison via KM-based QQ regression.
  Weibull ($R^2{=}0.994$, $\hat{k}{=}0.92$),
  Log-Normal ($R^2{=}0.998$, $\hat{\sigma}{=}2.88$), and
  Logistic ($R^2{=}0.994$, $\hat{s}{=}1.08$).
  Log-Normal achieves the highest $R^2$, consistent with its
  superior ranking metrics (Table~\ref{tab:shape_feat_ablation}).}
  \label{fig:dist_comparison}
\end{figure}

\paragraph{Distribution fitting.}
We fit Weibull, Log-Normal, and Logistic distributions to the
Kaplan--Meier (KM)~\cite{kaplan1958nonparametric} survival function
via quantile--quantile (QQ) regression (Figure~\ref{fig:dist_comparison}).
Log-Normal achieves the best fit ($R^2 = 0.998$),
followed by Logistic ($R^2 = 0.994$) and Weibull ($R^2 = 0.994$).
The Weibull QQ slope yields $\hat{k}_\text{QQ} = 0.919$,
confirming $k < 1$ (a slightly decreasing hazard).
The finding is consistent across customer segments:
light shoppers ($k = 0.948$), medium ($k = 0.930$), and
heavy buyers ($k = 0.908$) all exhibit $k < 1$.

A Weibull profile log-likelihood analysis
(Figure~\ref{fig:hazard}, right), maximizing over $\lambda$
for each $k$, yields $\hat{k}_\text{MLE} = 0.911$, which is consistent
with the QQ estimate.
Marginal scale estimation via profile likelihood is tractable
for Weibull thanks to its power-law form (closed-form concentration
of $\lambda$); for Log-Normal and Logistic, the fact that roughly
96.5\% of pairs are censored at a single time point renders the
marginal likelihood uninformative about the scale parameter, so we
determine optimal $\sigma$ by grid search
(Table~\ref{tab:shape_feat_ablation}).

\subsubsection{\textbf{Shape and Feature Ablation}}
\label{sec:shape_ablation}

Table~\ref{tab:shape_feat_ablation} presents a joint ablation of the
Weibull shape parameter $k$ and the input feature set.
All runs use depth 6, $\lambda{=}2.5$, LR 0.25, and share the same
train/test partition to isolate the effect of shape and features
from sampling noise. Here are the main observations:

\begin{table}[t]
\centering
\caption{Joint shape, feature, and distribution ablation.
All runs use depth 6, $\lambda{=}2.5$, and the same partition.
The first two blocks vary Weibull's shape parameter $k$ and the
feature set; the third block fixes $k{=}0.5$ (production features)
and swaps the AFT distribution family.}
\label{tab:shape_feat_ablation}
\footnotesize
\setlength{\tabcolsep}{3pt}
\begin{tabular}{lccccc}
\toprule
Config & $k$ & Feats & P@7d & P@14d & P@28d \\
\midrule
\multicolumn{6}{l}{\textit{Shape ablation (Weibull, production features)}} \\
\quad $k{=}0.5$ (decreasing) & 0.5 & 84 & \textbf{.3454} & \textbf{.3785} & \textbf{.4070} \\
\quad $k{=}1.0$ (exponential) & 1.0 & 84 & .3451 & .3783 & .4066 \\
\quad $k{=}2.0$ (increasing) & 2.0 & 84 & .3448 & .3777 & .4062 \\
\midrule
\multicolumn{6}{l}{\textit{Feature ablation (Weibull, +6 engineered timing features)}} \\
\quad $k{=}0.5$ & 0.5 & 90 & .3451 & .3784 & .4066 \\
\quad $k{=}1.0$ & 1.0 & 90 & .3452 & .3783 & .4065 \\
\quad $k{=}2.0$ (increasing) & 2.0 & 90 & .3448 & .3777 & .4057 \\
\midrule
\multicolumn{6}{l}{\textit{Distribution ablation (production features, $\sigma{=}2$)}} \\
\quad Weibull ($k{=}0.5$) & 0.5 & 84 & .3454 & .3785 & .4070 \\
\quad Log-Normal & --- & 84 & \textbf{.3461} & \textbf{.3788} & .4069 \\
\quad Logistic & --- & 84 & .3455 & .3787 & \textbf{.4071} \\
\bottomrule
\end{tabular}
\end{table}

\begin{itemize}
  \item{Shape matters more than hand-engineered features.}
Within the production feature set, moving the shape prior from
$k{=}1$ to $k{=}0.5$ lifts P@14d by roughly $0.05\%$.
Adding six engineered timing features (IPI-overdue indicators,
Weibull-derived CDF scores) does \emph{not} improve on the
production feature set; the best overall configuration uses the
production features alone.
This suggests the survival objective already extracts the timing
signal from the raw recency and frequency features, making explicit
timing engineering largely redundant.

  \item{Distribution choice: Log-Normal edges Weibull.}
Replacing the Weibull loss with Log-Normal produces the best overall
numbers (P@7d~$=~0.3461$, P@14d~$=~0.3788$), marginally exceeding
Weibull with $k{=}0.5$.
Logistic ties Weibull at 14 days and wins at 28 days.
Crucially, all three distributions outperform every per-horizon baseline, suggesting that the improvement is driven primarily by the AFT objective rather than by any particular distributional assumption.

  \item{Conditional-fit paradox.}
To understand \emph{why} Log-Normal outperforms Weibull despite the
data's Weibull-like marginal hazard, we perform a censoring-aware
conditional fit on each model's residuals
$r_i = \ln y_i - f(\mathbf{x}_i)$.
Figure~\ref{fig:cond_fit} (left) shows that the Gumbel distribution
fits residuals best for \emph{all three} training distributions,
not just the Weibull-trained model, yet Log-Normal wins on ranking
metrics (right panel).
The gap between Gumbel and Normal fits is smallest for the
Log-Normal model ($\Delta\text{LL}{=}908$) and largest for the
Logistic model ($\Delta\text{LL}{=}6{,}771$), suggesting that
Log-Normal's gradient landscape encourages better tree splits while
keeping residual structure close to Gumbel.
The loss function thus acts as both a fitting criterion and an
inductive bias: the bias that maximizes ranking performance need not
coincide with the one that best fits the conditional distribution.

\begin{figure}[t]
  \centering
  \includegraphics[width=\columnwidth]{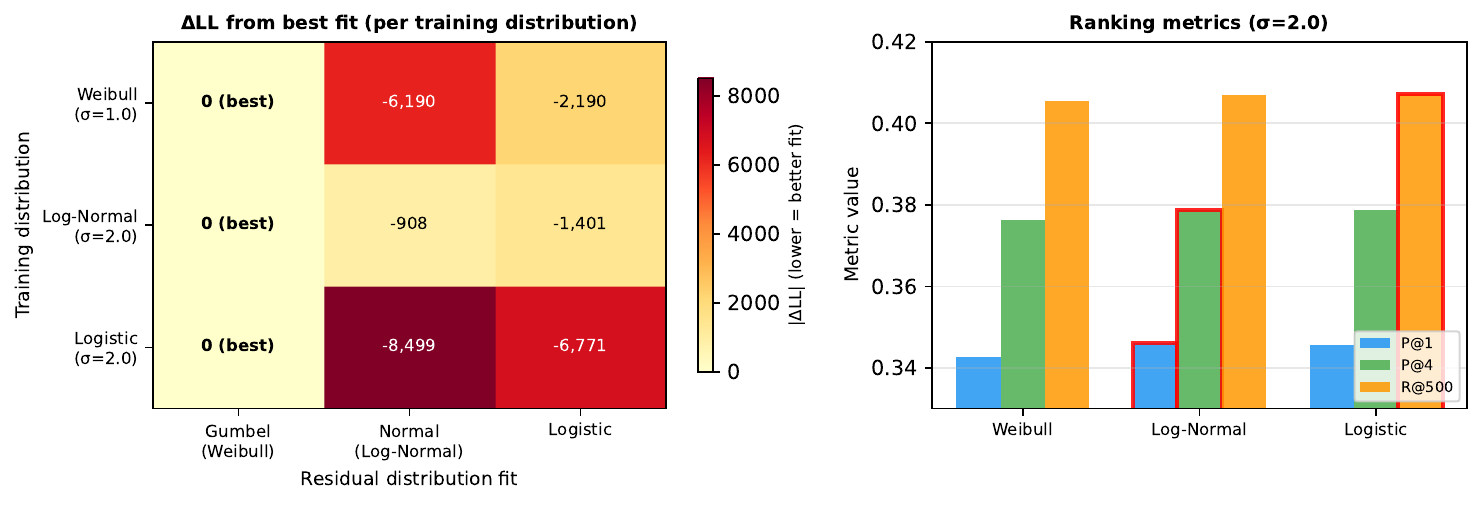}
  \caption{Conditional residual fit vs.\ ranking metrics.
  \textbf{Left}: $\Delta$LL heatmap. Gumbel fits residuals best
  across all training distributions; the gap is smallest for
  Log-Normal (908) vs.\ Logistic (6,771).
  \textbf{Right}: Ranking metrics favor Log-Normal despite
  Weibull's superior conditional fit.}
  \label{fig:cond_fit}
\end{figure}

  \item{Shape prior explains ranking differences.}
Among the Weibull models, $k{=}0.5$ beats $k{=}2.0$ at every
horizon (+0.19\% at P@14d, +0.22\% at P@28d), and ranking improves
monotonically as $k$ decreases from 2.0 to 0.5.
We lack intermediate grid points to say whether the true optimum
lies exactly at the empirical $\hat{k}_\text{MLE}{=}0.911$ or
somewhat below it.

  \item{A counterintuitive implication.}
The intuition that repurchase hazard should increase as items become \emph{more} overdue, is not supported by the population-level analysis.
The decreasing hazard reflects \emph{heterogeneity}: items that
remain non-repurchased after twenty days are disproportionately
long-cadence items (monthly detergent rather than weekly milk),
shifting the remaining at-risk population toward lower hazard.
XGBoost's feature-based model already captures per-item cadence,
so a slightly decreasing hazard prior avoids fighting the features.
\end{itemize}
\subsubsection{Failure Modes}

\paragraph{Never drop censored items.}
Three experiments filtered out right-censored items with large
lapsed-day values ($>$180 or $>$240 days).
All three collapsed catastrophically, with train--test gaps
exceeding 0.50, and even aggressive regularization
($\lambda{=}5$, subsample 0.8, LR 0.1, 800 rounds) could not
recover them.
Censored items anchor the survival tail; removing them destroys
the model's ability to generalize to long-cadence buyers.

\subsection{DT Ablations}
\label{sec:dt_ablations}

We evaluate five DT configurations that vary the number of intervals,
the feature set, and the training data volume
(Table~\ref{tab:dt_ablations}).
For these DT ablations we report fixed-$k$ recall (R@4, R@8
at 7 days); the cross-model comparison in
Table~\ref{tab:main} uses P@$h$.

\begin{table}[t]
\centering
\caption{DT ablation summary.
All configs: depth 6, LR 0.3, early stopping 20 rounds,
negative sample fraction 0.3.}
\label{tab:dt_ablations}
\footnotesize
\begin{tabular}{p{2.8cm}rrrr}
\toprule
Config & Trees & Feats & R@4 & R@8 \\
 & (actual) & & (7d) & (7d) \\
\midrule
3 intv, full data & 300 & 84 & 0.673 & 0.829 \\
3 intv + time-aware feats & 300 & 90 & 0.688 & 0.831 \\
3 intv, 800 max, half data & 225 & 84 & 0.686 & 0.832 \\
\textbf{3 intv + engagement feats, full} & \textbf{273} & \textbf{84} & \textbf{0.689} & \textbf{0.833} \\
4 intv + engagement feats, half data & 192 & 84 & 0.685 & 0.831 \\
\bottomrule
\end{tabular}
\end{table}

\paragraph{Three intervals are enough.}
Splitting the first week into $[0,3)$ and $[3,7)$ introduces a
fourth interval and inflates the training data by 31\%.
The model does notice the change, with \texttt{interval\_id} becoming
rank 7 by gain, yet Recall@4 at 7 days \emph{drops} from 0.689 to
0.685.
The 4-interval configuration is nevertheless informative in its
own right: the learned interval hazard rates are monotonically
increasing (Table~\ref{tab:hazard}).
This \emph{conditional} hazard, defined within each customer's
at-risk set, rises because items not yet repurchased become
progressively more overdue.
Crucially, it is fully compatible with the \emph{marginal} hazard
being decreasing ($\hat{k} < 1$, \S\ref{sec:empirical_hazard}):
at the population level, late survivors are disproportionately
long-cadence items, shifting the remaining at-risk set toward
lower hazard.

\begin{table}[t]
\centering
\footnotesize
\caption{Learned hazard rates by interval (4-interval DT config).
Monotonically increasing within each customer's at-risk set,
consistent with the AFT finding of $\hat{k} < 1$ at the
\emph{population} level (decreasing marginal hazard due to
heterogeneity).}
\label{tab:hazard}
\begin{tabular}{lrr}
\toprule
Interval & Mean $h(j)$ & Median $h(j)$ \\
\midrule
$[0, 3)$ days & 0.121 & 0.042 \\
$[3, 7)$ days & 0.195 & 0.080 \\
$[7, 14)$ days & 0.253 & 0.127 \\
$[14, 30)$ days & 0.340 & 0.232 \\
\bottomrule
\end{tabular}
\end{table}

\paragraph{Interval-relative features help where it matters.}
The six \texttt{*\_at\_end} features (Section~\ref{sec:dt})
simulate the customer's state at each interval boundary and improve
7- and 14-day recall by 1--2\%.
The strongest new signal is the population-level IPI gap at interval
end (rank 24 by gain), echoing the AFT finding that derived
timing-gap features outperform raw IPI values.

\paragraph{More data beats more trees.}
The full-data configuration (273 trees, early-stopped from 300)
outperforms the half-data configuration with an 800-tree cap
(225 trees actually used).
At this scale, additional data coverage buys more accuracy than
additional model capacity.

\subsection{Main Results}
\label{sec:main_results}

Table~\ref{tab:main} presents the head-to-head comparison against
production baselines.
All models are trained on the same customer partition and evaluated
with an identical label pipeline, so differences reflect model choice
rather than data or preprocessing artifacts.

\begin{table*}[t]
\centering
\caption{Main results: cross-horizon evaluation on a held-out
customer partition.
P@$h$ = precision-at-variable-$k$ at horizon $h$;
topk$N$ = fixed-$k$ precision at $k{=}N$ with 14-day labels.
$\dagger$ = production baseline.
$\ddagger$ = shape-aligned with the empirical $\hat{k}$.
Bold entries mark the best in each column.}
\label{tab:main}
\setlength{\tabcolsep}{3.5pt}
\small
\begin{tabular}{lrrrrrr}
\toprule
Model
  & P@7d & P@14d & P@28d
  & topk4 & topk8 & topk16 \\
\midrule
\multicolumn{7}{l}{\textit{Per-horizon binary classifiers (3 separate models, 84 feats)}} \\
\quad $F_\text{7d}$$^\dagger$
  & .3417 & .3717 & .3970
  & .5215 & .4486 & .3644 \\
\quad $F_\text{14d}$$^\dagger$
  & .3438 & .3757 & .4022
  & .5288 & .4540 & .3680 \\
\quad $F_\text{30d}$$^\dagger$
  & .3421 & .3758 & .4045
  & .5290 & .4539 & .3680 \\
\midrule
\multicolumn{7}{l}{\textit{DT-Hazard (single model, 84 feats, per-horizon ranking via $F(h)$)}} \\
\quad $F(7)$
  & .3406 & .3692 & .3944
  & .5151 & .4428 & .3603 \\
\quad $F(14)$
  & .3443 & .3748 & .4006
  & .5263 & .4515 & .3665 \\
\quad $F(30)$
  & .3445 & .3767 & .4041
  & .5305 & .4547 & .3686 \\
\midrule
\multicolumn{7}{l}{\textit{AFT Weibull (single model, 84 feats, horizon-invariant ranking)}} \\
\quad $k{=}2.0$ (increasing)
  & .3448 & .3777 & .4062
  & .5326 & .4566 & .3697 \\
\quad $k{=}1.0$ (exponential)$^\ddagger$
  & .3451 & .3783 & .4066
  & .5328 & .4568 & .3700 \\
\quad $k{=}0.5$ (decreasing)
  & .3454 & .3785 & .4070
  & .5330 & .4569 & .3701 \\
\midrule
\multicolumn{7}{l}{\textit{AFT alternate distributions (single model, 84 feats)}} \\
\quad Log-Normal ($\sigma{=}2$)
  & \textbf{.3461} & \textbf{.3788} & .4069
  & .5329 & \textbf{.4570} & \textbf{.3704} \\
\quad Logistic ($s{=}2$)
  & .3455 & .3787 & \textbf{.4071}
  & .5326 & .4569 & .3702 \\
\bottomrule
\end{tabular}
\end{table*}

The headline finding is simple: \textbf{a single survival model beats
three independently-trained per-horizon baselines at every horizon
we evaluate}.

\begin{itemize}
  \item \textbf{AFT wins overall.} 
The best AFT configuration (Log-Normal, 84 features) reaches
P@14d~$=~0.3788$, an improvement of $+0.82\%$ over the strongest
per-horizon baseline ($F_\text{14d}$, 0.3757).
Within the Weibull family, ranking improves monotonically as $k$
decreases from 2.0 to 0.5, confirming that the empirical hazard
analysis (\S\ref{sec:empirical_hazard}) translates directly into
ranking gains.

  \item \textbf{Survival models are more efficient.}
The production baseline trains three 800-tree models (2{,}400 trees
in aggregate).
The best AFT model uses a single ensemble of 330--765 trees, and
the DT model uses 273 trees.
Overall, a single survival model replaces roughly 2{,}400 trees with
$\sim$700. This is a $3\times$ reduction in training and serving compute
with no loss in ranking quality.

  \item \textbf{DT ties AFT at 7 days.}
The DT model ties AFT for 7-day precision (0.3445), but its best
score is achieved by the full CDF $F(30\text{d})$, which behaves
effectively as a horizon-invariant ranking.
This observation reinforces AFT's inductive bias, suggesting horizon-invariant
ordering does most of the work.

  \item \textbf{Independent classifiers transfer poorly.}
The 7d-trained model performs worst at 28 days ($0.3970$ vs.\
$0.4045$ for the horizon-matched model), and, more surprisingly,
$F_\text{14d}$ outperforms $F_\text{7d}$ even at 7 days
($0.3438$ vs.\ $0.3417$), indicating that the 14-day label may provide a more informative training signal.

\item \textbf{DT's full CDF dominates its horizon-specific scores.}
$F(30\text{d})$ outperforms $F(7\text{d})$ and $F(14\text{d})$ at
\emph{all} horizons, including 7 days.
Because $F(30\text{d}) = 1 - \prod_j(1-h_j)$ aggregates information
from every interval while $F(7\text{d}) = h_0$ uses only the first,
the richer signal yields better rankings even at short horizons.
For ranking, then, the practical takeaway is to always use the full
CDF; individual interval hazards are best reserved for
probability estimation.

\end{itemize}
\section{Feature Importance Under Survival Objectives}
\label{sec:feature_importance}

Switching from binary logistic to a survival objective visibly
reshuffles feature importance, even when the feature set is held
fixed.
Both the production classifier and our best AFT models (Weibull
$k{=}0.5$ and Log-Normal $\sigma{=}2$) use the same 84 production
features, and all importance values below are XGBoost's average gain
per split, computed on the trees actually used at
inference.\footnote{Both AFT models triggered early stopping;
importance is computed on the ``best-iteration'' slice of each
booster.}

The top three features \emph{total grocery orders}, \emph{item unavailability},
and \emph{base propensity score}, are stable across all three
objectives, and rank correlation with the
production model is high overall
(Spearman $\rho\!\approx\!0.87$--$0.88$ across the $\sim$80 shared
features).
Yet almost a quarter of the features shift by ten ranks or more, and
the pattern of those shifts is highly informative.

\paragraph{Features that rise under the survival objective.}
Channel-mix features, both at the customer level and at the
product-type level, become sharply more important.
The single largest riser is a \emph{product-type-level online-channel
order count}, which jumps roughly forty ranks under Weibull and
nearly fifty ranks under Log-Normal, moving from well outside the
production top 30 into the top 20 of both AFT models.
A \emph{customer-level online-channel share} climbs from rank~34 to
rank~10 (Weibull) and rank~\textbf{7} (Log-Normal), and
\emph{total online-channel orders} rises from rank~18 into the top 6
of both models.
These features encode shopping \emph{cadence}: online delivery and
in-store pickup run on different weekly rhythms, and the timing
objective picks up on this structure.
\emph{Item unavailability} rises from rank~3 to rank~2 in both AFT
models: knowing that an item was recently out-of-stock directly
constrains \emph{when} a customer will retry.
Related coarse indicators, including partial-fulfillment and lapse-bucket
flags, also rise substantially under Log-Normal, each climbing
roughly thirty ranks into the top 30.

\paragraph{Features that fall.}
Aggregate rolling-window frequency counts uniformly drop.
The starkest example is the \emph{3-month order total}, which
collapses from rank~12 to rank~54 (Weibull) / 46 (Log-Normal). This is the largest single fall in either survival model.
Weekly and monthly rolling-average order counts fall 15--20 places,
and a product-type-level in-store order count drops from rank~7 to
rank~24 / 26.
The survival objective demotes ``how much?'' features in favour of
``how recently?'' and ``through which channel?'' features.

\paragraph{Timing features: Log-Normal demotes deep history.}
Log-Normal retains only the \emph{recent} recency
signals, including days since the last purchase (rank 8), the second-last
(rank 18), and the third-last (rank 19), while demoting deep
history: days since the fifth-last purchase drops from rank~11 to
rank~20 ($-9$) and the fourth-last from rank~16 to rank~25 ($-9$).
Weibull is more conservative here, keeping the fifth-last recency
signal at rank~11 in line with production, and this is one of
several places where the two AFT distributions carve up the timing
signal differently.

\paragraph{Where the two AFT distributions disagree.}
The clearest example is an \emph{IPI--lapse residual}, defined as
the customer's typical inter-purchase interval minus their current
elapsed time, which is rank~18 in the Weibull model but only rank~44
in the Log-Normal model.
The residual encodes ``how far past the typical inter-purchase gap
this customer is''. This quantity aligns with Weibull's monotonically
increasing conditional hazard (the further overdue, the more urgent)
but is far less useful under Log-Normal's non-monotone, bell-shaped
hazard.
Conversely, Log-Normal leans more heavily on product-type-level
channel features and coarse lapse indicators.
The two distributions thus embody slightly different notions of
``timing'': Weibull as an ``overdue score'' and Log-Normal as a
channel- and cadence-mixture signal.

\paragraph{Personal vs.\ population IPI.}
A \emph{personal item IPI} (the customer's own median inter-purchase
interval for the item) sits at rank~15 under Log-Normal and rank~20
under Weibull, while its \emph{population-level} counterpart sits
just outside the top~30, functioning as a low-gain cold-start
routing feature.

\section{Calibration}
\label{sec:calibration}

Raw AFT outputs are not directly calibrated as repurchase
probabilities which are required by downstream applications.
For example, a typical predicted median $\hat{\lambda} \approx 5{,}700$ days, yields a raw $F(14) \approx 6 \times 10^{-6}$, which is orders of
magnitude below the empirical 14-day repurchase rate of $\sim$0.15.
Calibration is therefore necessary regardless of the choice of
distribution family.

\subsection{Distribution-Matched Parametric Calibration}

Each AFT distribution admits a natural calibration link that
preserves its parametric form.
We fit a shape parameter $a$ shared across horizons together with
per-horizon intercepts $b_t$ for $t \in \{7, 14, 28\}$:

\vspace{2pt}
\noindent\textit{Weibull (cloglog)}:
\begin{equation}
P_\text{calib}(T \le t) = 1 - \exp\!\bigl(-\exp(
  a \cdot \ln(t / \hat{\lambda}) + b_t)\bigr)
\label{eq:cloglog}
\end{equation}

\noindent\textit{Log-Normal (probit)}:
\begin{equation}
P_\text{calib}(T \le t) = \Phi\!\bigl(
  a \cdot \ln(t / \hat{\lambda}) + b_t\bigr)
\label{eq:probit}
\end{equation}

Both links share the same structure: a linear function of
$\ln(t/\hat{\lambda})$ with slope $a$ and per-horizon intercept
$b_t$.
The shared-$a$ formulation guarantees cross-horizon monotonicity by
construction, in contrast to isotonic calibration, which would
require per-row clipping.
Only four parameters are needed to cover three horizons.

\paragraph{Fitting.}
$(a, \{b_t\})$ are fit by minimizing joint per-horizon binary
cross-entropy on a held-out validation set via Nelder--Mead
optimization ($\sim$300 function evaluations, well under a second).

\subsection{Results}

We calibrate three AFT variants on the full test set
(hundreds-of-millions of customer--item pairs):
Exponential (Weibull $k{=}1$), Weibull $k{=}0.5$, and
Log-Normal $\sigma{=}2$.
The fitted parameters:

\vspace{2pt}
\noindent\textit{Exponential (Weibull $k{=}1$) + cloglog}:
$a = 0.998$, $b_{7} = 0.169$, $b_{14} = 0.134$, $b_{28} = 0.029$.

\noindent\textit{Weibull $k{=}0.5$ + cloglog}:
$a = 0.528$, $b_{7} = -0.319$, $b_{14} = -0.036$, $b_{28} = 0.177$.

\noindent\textit{Log-Normal $\sigma{=}2$ + probit}:
$a = 0.499$, $b_{7} = 0.100$, $b_{14} = 0.079$, $b_{28} = 0.036$.

\noindent
In all three cases, the fitted $a$ is close to $\approx 1/\sigma$, the inverse of the training scale parameter. The expected values are $1$ for Exponential,
$\approx 0.5$ for Weibull $k{=}0.5$ and Log-Normal $\sigma{=}2$). This agreement suggests that the calibration recovers the expected scale adjustment.
Cross-horizon monotonicity ($P(T \le 7) \le P(T \le 14) \le P(T \le 28)$)
holds with \textbf{zero violations} across every row of the
hundreds-of-millions holdout for all three calibrations, guaranteed
by the shared-$a$ formulation.

\begin{figure}[t]
  \centering
  \includegraphics[width=\columnwidth]{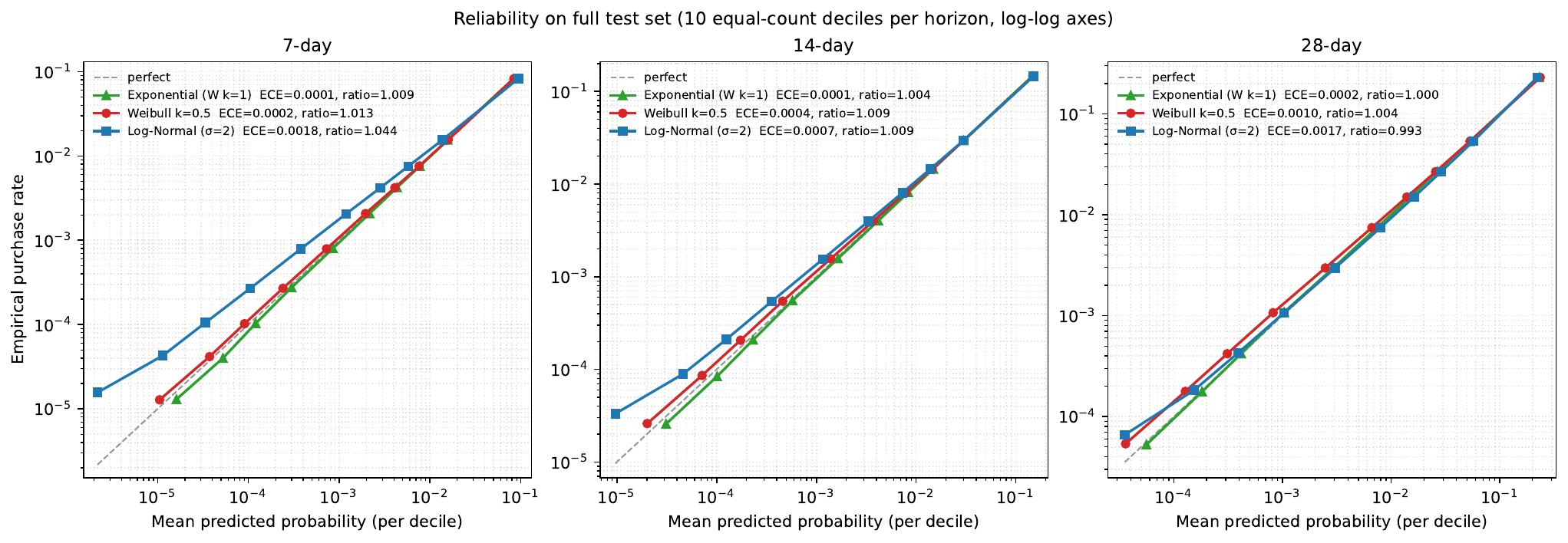}
  \caption{Reliability diagrams on the full test set
  (hundreds-of-millions of customer--item pairs,
  10 equal-count deciles per horizon,
  log--log axes to reveal the six decades of predicted probability the
  score distribution actually spans).
  Exponential (Weibull $k{=}1$, green) tracks the diagonal tightest
  across all three horizons; Weibull $k{=}0.5$ (red) is close
  behind; Log-Normal (blue) deviates most at the shortest horizon
  (7 days), where the bell-shaped hazard mismatches the empirical
  monotone-decreasing hazard the worst.}
  \label{fig:reliability_comparison}
\end{figure}

\paragraph{Calibration quality varies by an order of magnitude.}
Table~\ref{tab:calib_ops} reports ECE, sanity ratio
($\overline{P}/\overline{y}$), and top-tier operating points on
the 14-day horizon.
Exponential AFT is decisively the best-calibrated model:
its ECE at 14 days ($1.3\times10^{-4}$) is $\sim$3$\times$ lower
than Weibull $k{=}0.5$ ($3.6\times10^{-4}$) and $\sim$6$\times$
lower than Log-Normal ($7.5\times10^{-4}$), with sanity ratios
within $0.4\%$ of unity at every horizon.
The reliability diagrams (Figure~\ref{fig:reliability_comparison})
confirm that Exponential tracks the diagonal tightest across all
three horizons.

\paragraph{Why Log-Normal calibrates worse despite winning ranking.}
Log-Normal's ECE gap is not a fitting failure. The shared-$a$
fit converges to the joint negative log-likelihood (NLL) minimum
on the val set for all three distributions.
The gap traces to a hazard-shape mismatch.
Log-Normal has a non-monotone bell-shaped hazard: it rises then
falls, peaking near the mode of the underlying normal.
The empirical hazard (\S\ref{sec:empirical_hazard}) is
monotonically \emph{decreasing} ($\hat{k} \approx 0.9$), so
Log-Normal's rising left tail concentrates probability mass in
regions where the true hazard is already at its peak.
A single shared $a$ cannot correct this per-instance shape
mismatch, especially at the shortest horizon (7 days), where the
mismatch is largest.
Weibull's cloglog link, by contrast, is the exact CDF form of the
distribution the model was trained under; calibration only needs
to make small corrections.
Exponential ($k{=}1$) is the sweet spot: the flat hazard is
locally close to the empirical near-flat hazard, and the cloglog
link inherits Weibull's structural advantage.

\paragraph{Operating-point interpretation.}
The three variants also expose a coverage--precision structure at
the top confidence tier ($P{\ge}0.70$).
Exponential surfaces roughly $2\times$ the top-tier volume of
Weibull $k{=}0.5$ and $\sim$$3.6\times$ that of Log-Normal, at
81.6\% precision.
Weibull $k{=}0.5$ reaches 87.4\% and Log-Normal 90.5\%
precision but on a much smaller candidate set.
This is a direct consequence of calibration quality: Exponential's
$P{=}0.70$ scores land at empirical rate $\approx 0.82$
(mildly under-confident; $\overline{P}/\overline{y}$ close to 1
in the tier), whereas Log-Normal's $P{=}0.70$ scores over-shoot
substantially, which artificially inflates tier precision at the cost
of much lower coverage.
For most surfaces the Exponential trade-off dominates: honest
probabilities enable threshold-agnostic downstream use
(bidding, badge rendering, propensity fusion), while the raw
ranking remains within $0.3\%$ relative of the ranking-best
Log-Normal (\S\ref{sec:main_results}).

\begin{table}[t]
\centering
\footnotesize
\caption{Calibration comparison on the full test set
(14-day horizon; hundreds-of-millions of customer--item pairs).
``ratio'' is $\overline{P}/\overline{y}$; ideal $=1.000$.
``coverage'' is fraction of the full test set with $P{\ge}0.70$.
ECE computed over 10 equal-count deciles.}
\label{tab:calib_ops}
\begin{tabular}{lcccc}
\toprule
Calibration & ECE & ratio & P${\ge}$0.70 cov & P${\ge}$0.70 prec \\
\midrule
Exponential (Weibull $k{=}1$)
  & \textbf{0.00013} & \textbf{1.004} & \textbf{0.132\%} & 81.6\% \\
Weibull $k{=}0.5$
  & 0.00036 & 1.009 & 0.068\% & 87.4\% \\
Log-Normal ($\sigma{=}2$)
  & 0.00075 & 1.009 & 0.036\% & \textbf{90.5\%} \\
\bottomrule
\end{tabular}
\end{table}

\section{Discussion and Conclusion}
\label{sec:conclusion}

The core message of this paper is straightforward:
\textbf{asking ``when?'' instead of ``will they?'' yields a better
recommender essentially for free.}
A single AFT model replaces three per-horizon binary
classifiers, uses roughly $3\times$ fewer total trees, wins at every
horizon, and serves multiple web surfaces from a single ranked list.
The DT model offers per-horizon discrimination in principle, yet its
own best score is achieved by the full CDF, which behaves as a
horizon-invariant ranking, further validating AFT's inductive bias.

Along the way we observed several effects worth highlighting.
Distributional choice has a small but measurable impact: Log-Normal
edges Weibull and Logistic at shorter horizons, though all three
comfortably beat the per-horizon baselines.
Paradoxically, Weibull provides the tightest fit to conditional
residuals across all trained models, yet Log-Normal wins on ranking;
the loss function's role as an inductive bias matters more than its
role as a distributional fit.
The scale parameter does not change the ranking formula but does
affect learned predictions through the training-loss gradient, and
the Weibull shape prior matters under proper regularization.
One decision, however, is catastrophic: dropping censored items
destroys generalization to long-cadence buyers, no amount of
regularization recovers it.
Finally, the survival objective reshuffles feature importance in
predictable ways: channel-cadence and recent-recency signals rise
sharply while aggregate frequency counts fall, with direct
implications for feature engineering in any timing-sensitive
recommender.

On top of ranking, a 4-parameter parametric calibration turns the
raw survival CDF into deployable per-horizon probabilities.
Calibration quality separates the AFT family in a way ranking
does not: Exponential (Weibull $k{=}1$) reaches
ECE $\approx 10^{-4}$ with sanity ratios within $0.4\%$ of unity,
while Log-Normal, despite winning marginal fit and ranking, trails
by an order of magnitude on ECE because its bell-shaped hazard
mismatches the empirical monotone-decreasing hazard.
For probability-consuming surfaces (bidding, badge rendering,
propensity fusion) we adopt Exponential; for pure ranking,
Log-Normal remains the strongest.

\paragraph{Deployment efficiency.}
Beyond ranking quality, the survival formulation simplifies
operations: a single AFT model replaces three independently-trained
per-horizon classifiers, eliminating two-thirds of training
pipelines, serving endpoints, and A/B test configurations.
Feature engineering is shared across horizons, and the model's
ranking is reusable across web surfaces without re-inference. Only
the thin calibration layer changes per horizon.

\paragraph{Practical considerations.}
AFT ranking is horizon-invariant by construction; applications that
genuinely require different orderings per horizon should use the
DT formulation, at the cost of a $\sim 3\times$ data expansion
and negative sampling.
All results reported here are offline; online A/B testing on live
traffic is the natural next step.

\paragraph{Limitations.}
Our experiments are offline and drawn from a single, proprietary
grocery retailer's dataset; we have not yet validated the findings on
public repurchase benchmarks (e.g.,
Dunnhumby, Instacart, Tafeng) or on non-grocery verticals such as
health-and-beauty or general merchandise, where inter-purchase
cadences and censoring patterns may differ substantially.
Both the distribution family and its scale parameter are estimated
globally; per-category parameters (e.g., weekly milk vs.\ monthly
detergent) may yield further gains.
The AFT model does not model basket-level dependencies between
items bought together, and the 30-day censoring window truncates
long-cadence items whose true repurchase times extend beyond it.

\paragraph{Future work.}
Several avenues follow naturally.
Reproducing the empirical hazard analysis and the AFT-vs-DT
comparison on public repurchase benchmarks would test whether the
$\hat{k} < 1$ finding and the calibration--ranking trade-off hold
beyond a single retailer.
Per-item distribution parameters could better capture multi-modal
repurchase cycles; CDF-based urgency reranking could power
time-sensitive surfaces; joint training with basket size and order
timing could unify the timing stack; and survival scores could serve
as informative priors in generative retrieval pipelines.

\newpage

\bibliographystyle{ACM-Reference-Format}
\bibliography{refs}

@inproceedings{chen2016xgboost,
  title     = {{XGBoost}: A Scalable Tree Boosting System},
  author    = {Chen, Tianqi and Guestrin, Carlos},
  booktitle = {Proceedings of the 22nd ACM SIGKDD International Conference on Knowledge Discovery and Data Mining},
  pages     = {785--794},
  year      = {2016},
  publisher = {ACM},
  address   = {New York, NY, USA}
}

@article{cox1972regression,
  title   = {Regression Models and Life-Tables},
  author  = {Cox, D. R.},
  journal = {Journal of the Royal Statistical Society: Series B (Methodological)},
  volume  = {34},
  number  = {2},
  pages   = {187--202},
  year    = {1972}
}

@article{efron1988logistic,
  title   = {Logistic Regression, Survival Analysis, and the Kaplan-Meier Curve},
  author  = {Efron, Bradley},
  journal = {Journal of the American Statistical Association},
  volume  = {83},
  number  = {402},
  pages   = {414--425},
  year    = {1988}
}

@inproceedings{hidasi2016session,
  title     = {Session-Based Recommendations with Recurrent Neural Networks},
  author    = {Hidasi, Bal{\'a}zs and Karatzoglou, Alexandros and Baltrunas, Linas and Tikk, Domonkos},
  booktitle = {International Conference on Learning Representations (ICLR)},
  year      = {2016},
  publisher = {OpenReview.net},
  address   = {San Juan, Puerto Rico}
}

@inproceedings{sun2019bert4rec,
  title     = {{BERT4Rec}: Sequential Recommendation with Bidirectional Encoder Representations from Transformer},
  author    = {Sun, Fei and Liu, Jun and Wu, Jian and Pei, Changhua and Lin, Xiao and Ou, Wenwu and Jiang, Peng},
  booktitle = {Proceedings of the 28th ACM International Conference on Information and Knowledge Management},
  pages     = {1441--1450},
  year      = {2019},
  publisher = {ACM},
  address   = {New York, NY, USA}
}

@inproceedings{chandar2022survival,
  title     = {Using Survival Models to Estimate User Engagement in Online Experiments},
  author    = {Chandar, Praveen and St. Thomas, Brian and Maystre, Lucas and Pappu, Vijay and Sanchis-Ojeda, Roberto and Wu, Tiffany and Carterette, Ben and Lalmas, Mounia and Jebara, Tony},
  booktitle = {Proceedings of the ACM Web Conference 2022 (WWW)},
  pages     = {3186--3195},
  year      = {2022},
  publisher = {ACM},
  address   = {New York, NY, USA}
}

@inproceedings{kapoor2015just,
  title     = {Just in Time Recommendations: Modeling the Dynamics of Boredom in Activity Streams},
  author    = {Kapoor, Komal and Subbian, Karthik and Srivastava, Jaideep and Schrater, Paul},
  booktitle = {Proceedings of the Eighth ACM International Conference on Web Search and Data Mining},
  pages     = {233--242},
  year      = {2015},
  publisher = {ACM},
  address   = {New York, NY, USA}
}

@inproceedings{zhu2017next,
  title     = {What to Do Next: Modeling User Behaviors by Time-{LSTM}},
  author    = {Zhu, Yu and Li, Hao and Liao, Yikang and Wang, Beidou and Guan, Ziyu and Liu, Haifeng and Cai, Deng},
  booktitle = {Proceedings of the 26th International Joint Conference on Artificial Intelligence (IJCAI)},
  pages     = {3602--3608},
  year      = {2017},
  publisher = {AAAI Press},
  address   = {Melbourne, Australia}
}

@inproceedings{ma2018modeling,
  title     = {Modeling Task Relationships in Multi-Task Learning with Multi-Gate Mixture-of-Experts},
  author    = {Ma, Jiaqi and Zhao, Zhe and Yi, Xinyang and Chen, Jilin and Hong, Lichan and Chi, Ed H.},
  booktitle = {Proceedings of the 24th ACM SIGKDD International Conference on Knowledge Discovery and Data Mining},
  pages     = {1930--1939},
  year      = {2018},
  publisher = {ACM},
  address   = {New York, NY, USA}
}

@book{singer2003applied,
  title     = {Applied Longitudinal Data Analysis: Modeling Change and Event Occurrence},
  author    = {Singer, Judith D. and Willett, John B.},
  year      = {2003},
  publisher = {Oxford University Press},
  address   = {New York, NY, USA}
}

@inproceedings{wan2018representing,
  title     = {Representing and Recommending Shopping Baskets with Complementarity, Compatibility and Loyalty},
  author    = {Wan, Mengting and Wang, Di and Liu, Jie and Bennett, Paul and McAuley, Julian},
  booktitle = {Proceedings of the 27th ACM International Conference on Information and Knowledge Management},
  pages     = {1133--1142},
  year      = {2018},
  publisher = {ACM},
  address   = {New York, NY, USA}
}

@inproceedings{ariannezhad2022recanet,
  title     = {{ReCANet}: A Repeat Consumption-Aware Neural Network for Next Basket Recommendation in Grocery Shopping},
  author    = {Ariannezhad, Mozhdeh and Jullien, S{\'e}bastien and Li, Ming and Fang, Min and Schelter, Sebastian and de Rijke, Maarten},
  booktitle = {Proceedings of the 45th International ACM SIGIR Conference on Research and Development in Information Retrieval},
  pages     = {1240--1250},
  year      = {2022},
  publisher = {ACM},
  address   = {New York, NY, USA}
}

@article{kaplan1958nonparametric,
  title   = {Nonparametric Estimation from Incomplete Observations},
  author  = {Kaplan, E. L. and Meier, Paul},
  journal = {Journal of the American Statistical Association},
  volume  = {53},
  number  = {282},
  pages   = {457--481},
  year    = {1958}
}

@book{kalbfleisch2002statistical,
  title     = {The Statistical Analysis of Failure Time Data},
  author    = {Kalbfleisch, John D. and Prentice, Ross L.},
  edition   = {2nd},
  year      = {2002},
  publisher = {John Wiley \& Sons},
  address   = {Hoboken, NJ, USA}
}

@article{wei1992accelerated,
  title   = {The Accelerated Failure Time Model: A Useful Alternative to the {Cox} Regression Model in Survival Analysis},
  author  = {Wei, L. J.},
  journal = {Statistics in Medicine},
  volume  = {11},
  number  = {14--15},
  pages   = {1871--1879},
  year    = {1992}
}

@inproceedings{bhagat2018buyitagain,
  title     = {Buy It Again: Modeling Repeat Purchase Recommendations},
  author    = {Bhagat, Rahul and Muralidharan, Srevatsan and Lobzhanidze, Alex and Vishwanath, Shankar},
  booktitle = {Proceedings of the 24th ACM SIGKDD International Conference on Knowledge Discovery and Data Mining},
  pages     = {62--70},
  year      = {2018},
  publisher = {ACM},
  address   = {New York, NY, USA}
}

@article{allison1982discrete,
  title   = {Discrete-Time Methods for the Analysis of Event Histories},
  author  = {Allison, Paul D.},
  journal = {Sociological Methodology},
  volume  = {13},
  pages   = {61--98},
  year    = {1982}
}

@inproceedings{fpmc,
  title     = {Factorizing Personalized {Markov} Chains for Next-Basket Recommendation},
  author    = {Rendle, Steffen and Freudenthaler, Christoph and Schmidt-Thieme, Lars},
  booktitle = {Proceedings of the 19th International Conference on World Wide Web},
  pages     = {811--820},
  year      = {2010},
  publisher = {ACM},
  address   = {New York, NY, USA}
}

@inproceedings{yu2016dream,
  title     = {A Dynamic Recurrent Model for Next Basket Recommendation},
  author    = {Yu, Feng and Liu, Qiang and Wu, Shu and Wang, Liang and Tan, Tieniu},
  booktitle = {Proceedings of the 39th International ACM SIGIR Conference on Research and Development in Information Retrieval},
  pages     = {729--732},
  year      = {2016},
  publisher = {ACM},
  address   = {New York, NY, USA}
}

@inproceedings{hu2019sets2sets,
  title     = {{Sets2Sets}: Learning from Sequential Sets with Neural Networks},
  author    = {Hu, Haoji and He, Xiangnan},
  booktitle = {Proceedings of the 25th ACM SIGKDD International Conference on Knowledge Discovery and Data Mining},
  pages     = {1491--1499},
  year      = {2019},
  publisher = {ACM},
  address   = {New York, NY, USA}
}

@inproceedings{le2019beacon,
  title     = {Correlation-Sensitive Next-Basket Recommendation},
  author    = {Le, Duc-Trong and Lauw, Hady W. and Fang, Yuan},
  booktitle = {Proceedings of the 28th International Joint Conference on Artificial Intelligence (IJCAI)},
  pages     = {2808--2814},
  year      = {2019},
  publisher = {IJCAI},
  address   = {Macao, China}
}

@inproceedings{yu2020dnntsp,
  title     = {Predicting Temporal Sets with Deep Neural Networks},
  author    = {Yu, Le and Sun, Leilei and Du, Bowen and Liu, Chuanren and Xiong, Hui and Lv, Weifeng},
  booktitle = {Proceedings of the 26th ACM SIGKDD International Conference on Knowledge Discovery and Data Mining},
  pages     = {1083--1091},
  year      = {2020},
  publisher = {ACM},
  address   = {New York, NY, USA}
}

@inproceedings{hu2020tifuknn,
  title     = {Modeling Personalized Item Frequency Information for Next-Basket Recommendation},
  author    = {Hu, Haoji and He, Xiangnan and Gao, Jinyang and Zhang, Zhi-Li},
  booktitle = {Proceedings of the 43rd International ACM SIGIR Conference on Research and Development in Information Retrieval},
  pages     = {1071--1080},
  year      = {2020},
  publisher = {ACM},
  address   = {New York, NY, USA}
}

@inproceedings{ren2019repeatnet,
  title     = {{RepeatNet}: A Repeat Aware Neural Recommendation Machine for Session-Based Recommendation},
  author    = {Ren, Pengjie and Chen, Zhumin and Li, Jing and Ren, Zhaochun and Ma, Jun and de Rijke, Maarten},
  booktitle = {Proceedings of the AAAI Conference on Artificial Intelligence},
  volume    = {33},
  pages     = {4806--4813},
  year      = {2019},
  publisher = {AAAI Press},
  address   = {Palo Alto, CA, USA}
}

@article{li2023realitycheck,
  title     = {A Next Basket Recommendation Reality Check},
  author    = {Li, Ming and Jullien, Sami and Ariannezhad, Mozhdeh and de Rijke, Maarten},
  journal   = {ACM Transactions on Information Systems},
  volume    = {41},
  number    = {4},
  pages     = {116:1--116:29},
  year      = {2023},
  publisher = {ACM}
}

@inproceedings{case,
  title     = {{CASE}: Cadence-Aware Set Encoding for Large-Scale Next Basket Repurchase Recommendation},
  author    = {Cao, Yanan and Ranjan, Ashish and Subramaniam, Sinduja and Korpeoglu, Evren and Nag, Kaushiki and Achan, Kannan},
  booktitle = {Proceedings of the 49th International ACM SIGIR Conference on Research and Development in Information Retrieval (Industry Track)},
  year      = {2026},
  publisher = {ACM},
  doi       = {10.48550/arXiv.2604.06718}
}

@inproceedings{dacrema2019,
  title     = {Are We Really Making Much Progress? {A} Worrying Analysis of Recent Neural Recommendation Approaches},
  author    = {Ferrari Dacrema, Maurizio and Cremonesi, Paolo and Jannach, Dietmar},
  booktitle = {Proceedings of the 13th ACM Conference on Recommender Systems},
  pages     = {101--109},
  year      = {2019},
  publisher = {ACM},
  address   = {New York, NY, USA}
}

@inproceedings{rendle2020neural,
  title     = {Neural Collaborative Filtering vs.\ Matrix Factorization Revisited},
  author    = {Rendle, Steffen and Krichene, Walid and Zhang, Li and Anderson, John},
  booktitle = {Proceedings of the 14th ACM Conference on Recommender Systems},
  pages     = {240--248},
  year      = {2020},
  publisher = {ACM},
  address   = {New York, NY, USA}
}

@article{shwartzziv2022tabular,
  title     = {Tabular Data: Deep Learning is Not All You Need},
  author    = {Shwartz-Ziv, Ravid and Armon, Amitai},
  journal   = {Information Fusion},
  volume    = {81},
  pages     = {84--90},
  year      = {2022},
  publisher = {Elsevier}
}

@inproceedings{grinsztajn2022why,
  title     = {Why Do Tree-Based Models Still Outperform Deep Learning on Typical Tabular Data?},
  author    = {Grinsztajn, L{\'e}o and Oyallon, Edouard and Varoquaux, Ga{\"e}l},
  booktitle = {Advances in Neural Information Processing Systems 35: Datasets and Benchmarks Track},
  year      = {2022}
}

\appendix

\end{document}